\documentclass[10pt,twocolumn,letterpaper]{article}

\usepackage[pagenumbers]{cvpr} 

\newcommand{\red}[1]{{\color{red}#1}}

\usepackage{amsmath,amsfonts,bm}

\def\eqref#1{equation~\ref{#1}}

\def\1{\bm{1}}

\def\rvx{{\mathbf{x}}}

\DeclareMathAlphabet{\mathsfit}{\encodingdefault}{\sfdefault}{m}{sl}
\SetMathAlphabet{\mathsfit}{bold}{\encodingdefault}{\sfdefault}{bx}{n}

\newcommand{\KL}{D_{\mathrm{KL}}}

\usepackage{algorithm}
\usepackage{algpseudocode}

\usepackage[toc,page,header]{appendix}
\usepackage{minitoc}

\newtheorem{theorem}{Theorem}[section]

\newcommand{\se}[1]{{\color{red} {\small\bf SE:} #1}}

\newcommand{\alg}{\text{DDRL}\xspace}
\newcommand{\reft}{\text{ref}\xspace}

\definecolor{cvprblue}{rgb}{0.21,0.49,0.74}
\usepackage[pagebackref,breaklinks,colorlinks,allcolors=cvprblue]{hyperref}
\usepackage{multirow}

\def\paperID{} 
\def\confName{CVPR}
\def\confYear{2026}

\title{Data-regularized Reinforcement Learning for Diffusion Models at Scale}

\author{
\text{Haotian Ye}$^{12}$\thanks{Work done as an intern at NVIDIA, and used in \href{https://research.nvidia.com/labs/dir/cosmos-predict2.5/}{Cosmos-Predict2.5}. Code \href{https://github.com/nvidia-cosmos/cosmos-rl/blob/main/examples/ddrl.md}{here}. Correspond to: \url{haotianye@stanford.edu}. } ~ \text{Kaiwen Zheng}$^{23}$ ~ \text{Jiashu Xu}$^{2}$ ~ \text{Puheng Li}$^{1}$ ~ \text{Huayu Chen}$^{23}$ ~ \text{Jiaqi Han}$^{1}$ ~ \text{Sheng Liu}$^{1}$ \\
~ \text{Qinsheng Zhang}$^{2}$ ~ \text{Hanzi Mao}$^{2}$ ~ \text{Zekun Hao}$^{2}$ ~ \text{Prithvijit Chattopadhyay}$^{2}$ ~ \text{Dinghao Yang}$^{2}$ \\~ \text{Liang Feng}$^{2}$ 
~ \text{Maosheng Liao}$^{2}$ ~ \text{Junjie Bai}$^{2}$ ~ \text{Ming-Yu Liu}$^{2}$ ~ \text{James Zou}$^{1}$ ~ \text{Stefano Ermon}$^{1}$ \\
~\text{$^1$Stanford University} \quad
~\text{$^2$NVIDIA} \quad
~\text{$^3$Tsinghua University} \\
\url{https://research.nvidia.com/labs/dir/ddrl/}
}

\begin{document}
\maketitle
\begin{abstract}
Aligning generative diffusion models with human preferences via reinforcement learning (RL) is critical yet challenging. Most existing algorithms are often vulnerable to reward hacking, such as quality degradation, over-stylization, or reduced diversity. Our analysis demonstrates that this can be attributed to the inherent limitations of their regularization, which provides unreliable penalties. We introduce Data-regularized Diffusion Reinforcement Learning (DDRL), a novel framework that uses the \textit{forward} KL divergence to anchor the policy to an off-policy data distribution. Theoretically, DDRL enables robust, unbiased integration of RL with standard diffusion training. Empirically, this translates into a simple yet effective algorithm that combines reward maximization with diffusion loss minimization. With over a million GPU hours of experiments and ten thousand double-blind human evaluations, we demonstrate on high-resolution video generation tasks that DDRL significantly improves rewards while alleviating the reward hacking seen in baselines, achieving the highest human preference and establishing a robust and scalable paradigm for diffusion post-training.

\vspace{-10pt}
\end{abstract}

\section{Introduction}
\begin{figure*}[!t]
    \centering
    \includegraphics[width=0.98\linewidth]{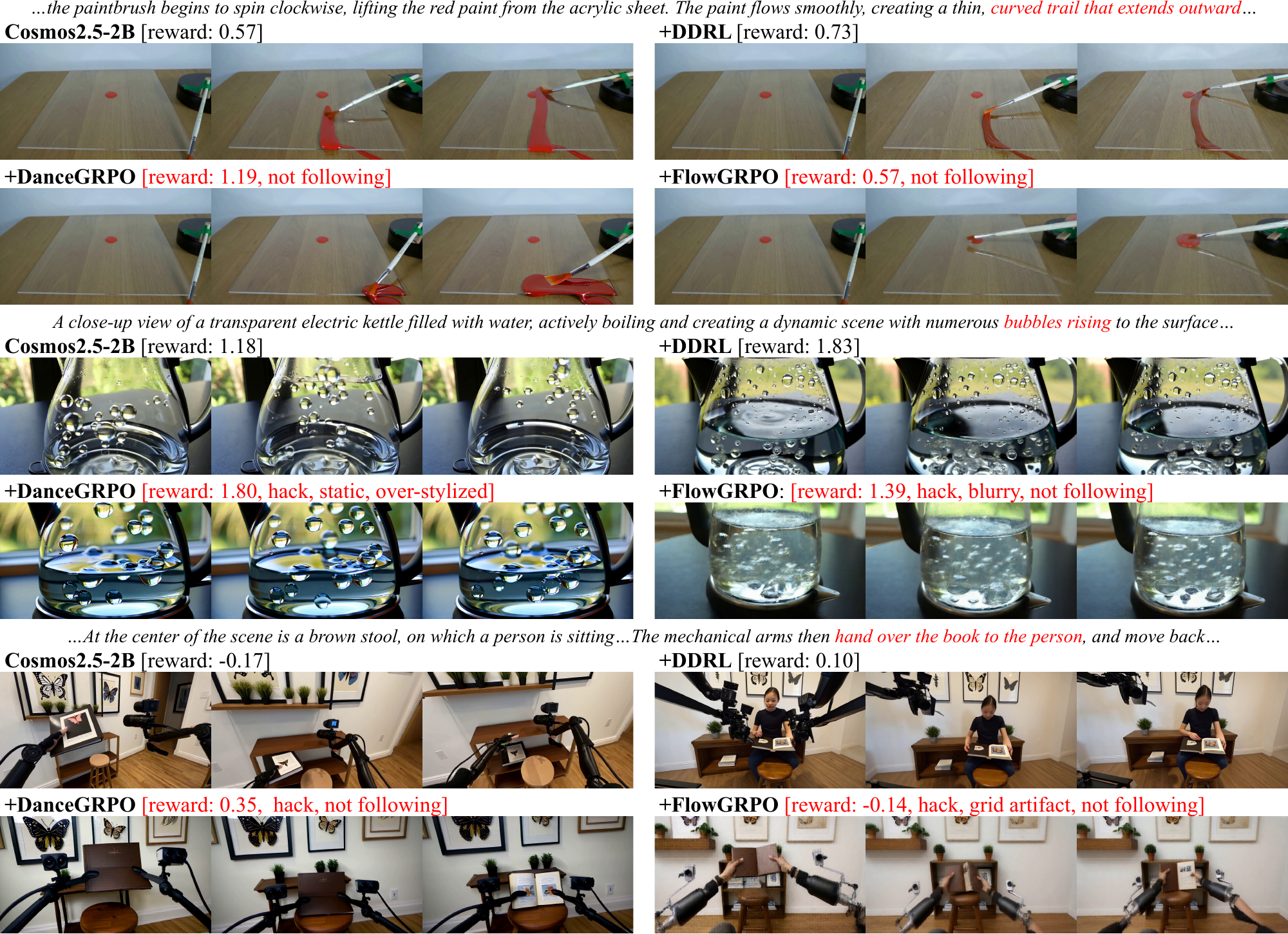}
    \vspace{-10pt}
    \caption{Generated videos after RL with different algorithms from the same checkpoint. \alg improves rewards by generating prompt-aligned and realistic videos, while DanceGRPO and FlowGRPO increase rewards with over-colorized, over-stylized, and unrealistic videos. }
    \label{fig:video_hacking}
    \vspace{-10pt}
\end{figure*}

Recent developments in 
generative models, including denoising diffusion models \cite{nichol2021improved,song2020score,song2020denoising,esser2024scaling} and the flow-based counterparts~\cite{lipman2022flow,liu2022flow,song2023consistency}, have demonstrated their remarkable capabilities in high-quality images and videos generation~\cite{agarwal2025cosmos, ali2025world}. 
With impressive scalability to billions of data points~\cite{schuhmann2022laion}, diffusion models have the potential to serve as foundational generative models across a wide range of applications~\cite{song2019generative,nichol2021improved,rombach2022high,ye2024tfg}.
To align these models with human preferences, reinforcement learning (RL) has emerged as a powerful post-training paradigm.
Algorithms like RLHF and GRPO~\cite{achiam2023gpt,ouyang2022training,shao2024deepseekmath} post-train a generative model $p_\theta$ from a reference model $p_\reft$ to maximize the rewards that represent human preferences, while regularizing the \textit{reverse} KL divergence $\KL(p_\theta\|p_\reft)$ to control their difference.
Thereafter, they have been adopted to post-train diffusion models as well~\cite{black2023training,wallace2024diffusion,liu2025flow,xue2025dancegrpo,gao2025seedance}. 



While critical, applying RL to diffusion models at scale is challenging. 
Most vision reward models are learned from human preferences data and are non-verifiable: they accurately reflect human preferences only in a neighborhood of the training data manifold. 
As $p_\theta$ is optimized for reward maximization, its multi-step, Markovian sampling procedure can easily exploit out-of-distribution (OOD) regions away from the data manifold, where the reference model is barely trained and fails to provide accurate regularization. 
This issue lies in the adoption of on-policy samples for regularization, and makes existing algorithms vulnerable to reward hacking~\cite{skalse2022defining,zhai2025mira}, a phenomenon where models generate human-unpreferred samples that nevertheless receive high rewards, resulting in issues like diversity decline, quality degradation, and over-stylization. 
We illustrate in \cref{fig:video_hacking} the low-quality, unrealistic video generations that nevertheless achieve a high reward, despite the KL divergence remaining controlled (\cref{fig:largebeta}). Similarly, \cref{fig:t2i} illustrates the over-simplified images generated by existing methods.
We are motivated by the fundamental question: \textit{can we design a more principled RL algorithm for diffusion models that is effective and inherently robust to hacking?} 
To address the issue caused by on-policy regularization, we introduce \textbf{Data-regularized Diffusion Reinforcement Learning} (\alg), a novel framework that anchors to off-policy data distributions. Intuitively, \alg maximizes relative rewards while regularizing the \textit{diffusion loss} on a dataset that is either real or synthetic to prevent unreasonable generations. Theoretically, \alg stems from the \textit{forward} KL divergence, which is equivalent to the diffusion training objective and guarantees unbiased RL optimization. It is robust to reward hacking and seamlessly integrates RL with supervised learning (SFT), scaling effectively by balancing on-policy exploration with off-policy data quality.
We conduct extensive experiments on large-scale video and image generation tasks, utilizing over one million H100 GPU hours and ten thousand double-blind human voting results. Under all settings with different rewards, base models, and conditions, \alg consistently improves rewards while also maintaining the highest human voting preferences, while existing baselines either fail to increase rewards (with large regularization and early stopping) or provide hacked generations that are consistently not favored by voters. 
Our empirical analysis shows that they hack by trading off one reward score with another, while \alg can bring Pareto improvement. Furthermore, we ablate \alg and demonstrate its potential to integrate SFT and RL into a single stage, achieving comparable quality with significantly higher data efficiency. We also justify the effectiveness of \alg when using synthetic data for regularization.

In summary, we identify on-policy regularization as the primary cause of reward hacking in diffusion RL, and propose \alg that scalably achieves reward improvements and human preference with off-policy data regularization. With its theoretical guarantee, elegant integration of post-training, as well as empirical simplicity and effectiveness, \alg establishes a principled framework for future diffusion reinforcement learning at scale.

\section{Backgrounds \& Preliminaries}
\label{sec:preliminaries}


\paragraph{Diffusion generative models} are a class of generative models that learn a conditional data distribution $p_{\text{data}}(\rvx|c)$ by reversing a fixed noise-adding process\footnote{
For simplicity, we formulate the framework and our algorithm for diffusion models, while it is directly applicable to other variants.}, where $c$ is a given condition~\cite{nichol2021improved,song2020score,song2020denoising,esser2024scaling}. 
Given a noise schedule $\{\beta_t\}_{t=1}^T \in (0, 1)$, the forward process is defined as \begin{align}
q(\rvx_0 |c ) &= p_\text{data}(\rvx|c), \notag \\
    q(\rvx_t | \rvx_{t-1}) &= \mathcal{N}(\sqrt{1-\beta_t} \rvx_{t-1}, \beta_t \mathbf{I}).\label{eq:add_noise}
\end{align}
A diffusion model $\boldsymbol \epsilon_\theta$ learns the distribution by 
\begin{equation}
    p_\theta(\rvx_{0:T}|c) = p_\theta(\rvx_T) \prod_{t=1}^T p_\theta(\rvx_{t-1}|\rvx_t, c),
\end{equation}
where $\rvx _T \sim \mathcal{N}(\mathbf{0}, \mathbf{I})$ and $\rvx_{t-1} \sim \mathcal N(\boldsymbol {\mu}_\theta(\rvx_t, t, c) ,\sigma_t^2 \mathbf I)$.
Here $\boldsymbol {\mu}_\theta$ can be computed from $\rvx_t$ and the model-predicted noise $\boldsymbol{\epsilon}_\theta(\rvx_t, t, c)$. 
The goal is to minimize the KL divergence between $q(\rvx_{0:T}|c)$ and $p_\theta(\rvx_{0:T}|c)$, i.e. 
\begin{align}
\KL(q(\cdot|c) \| p_\theta(\cdot|c)) \triangleq \int_{\rvx_{0:T}} q(\rvx_{0:T}|c) \log \frac{q(\rvx_{0:T}|c)}{p_\theta(\rvx_{0:T}|c)} \mathrm d \rvx. \label{eq:diffusion_KL}
\end{align}
To this end, the model is trained by minimizing:
\begin{align}
    \mathcal L(\theta; p_\text{data}) = \mathbb E_{t, \rvx_t \sim q(\rvx_t|c)} \big[w_t\|\boldsymbol{\epsilon}_\theta(\rvx_t,t,c) - \boldsymbol{\epsilon} \|^2\big],\label{eq:diffusion_loss}
\end{align}
where $q$ is constructed from $p_\text{data}$ as defined above, $\boldsymbol{\epsilon} $ is the Gaussian noise used to sample $\rvx_t$ from $q$, and $w_t$ is $\theta$-independent coefficient calculated from $\beta_t$.
As a classical result, \cref{eq:diffusion_loss} can be proven equivalent to \cref{eq:diffusion_KL} up to a $\theta$-independent constant, and thus both objectives share the same optimum at which $p_\theta(\rvx_0|c)$ equals $p_\text{data}(\rvx|c)$.

\paragraph{Reward Models.}Similar to RL for language models~\citep{ouyang2022training,shao2024deepseekmath,rafailov2023direct}, we are given a reward model $r(\rvx_0, c)$, which represents the user's preference to a sample $\rvx_0$. Remarkably, $r$ can be any rule-based function or neural network, and it can represent a variety of properties such as a video's smoothness, an image's physical reasonability, as well as a sample's prompt adherence~\cite{liu2025flow,xue2025dancegrpo,wallace2024diffusion}. 
In the vision domain, most rewards are \textit{non-verifiable}, i.e., they are an approximation of the underlying true rewards $r^*(\rvx_0,c)$ that we have no access to, and 
are only accurate in a neighborhood of the real data manifold. In other words, a higher reward does not necessarily imply a better sample. 


\paragraph{Reinforcement learning for diffusion.} 
The objective is to post-train $p_\theta$ to generate samples from
\begin{align}
    p_\theta(\rvx_0|c) \propto p_{\text{data}}(\rvx_0|c) \exp\left(r(\rvx_0,c)/{\beta}\right),
    \label{eq:target_dist}    
\end{align}
where $\beta$ is a temperature parameter. 
Applying RL to diffusion models is not straightforward and reformulation is required.
Existing literature shows that the denoising process can be formulated as a Markov Decision Process (MDP)~\cite{black2023training}, making it possible to post-train the diffusion model via RL algorithms and reach \cref{eq:target_dist}. Specifically, the diffusion model $\boldsymbol{\epsilon}_\theta$ is the policy whose action is to sample $\rvx_{t-1}$ from current state $(\rvx_t, t, c)$, reaching state $(\rvx_{t-1}, t-1, c)$. 
The reward $r(\rvx_0,c)$ is provided at the terminal step $t=0$. The diffusion policy is post-trained  by maximizing
\begin{align}
    \mathcal J_{\text{RL}}(p_\theta) = \mathbb E_{p_\theta(\rvx_0|c)}\Big[r(\rvx_0,c)/\beta \Big] -  \KL(p_\theta|| p_{\text{ref}}).
    \label{eq:rl_objective}
\end{align}
Notice that $p_\text{ref}(\rvx_{0:T}|c)$ in the formula typically represents a reference diffusion model that is trained on $p_\text{data}$. 
Classical RL theory show that its optimum satisfies $p_\theta(\rvx_0|c) \propto p_\text{ref}(\rvx_0 |c) \exp (r(\rvx_0,c) / \beta)$~\cite{sutton1998introduction}, which is an approximation of \cref{eq:target_dist} if $p_\text{ref}$ is close to $p_\text{data}$.

\paragraph{Optimization.} 
\cref{eq:rl_objective} leverages \textit{reverse} KL divergence to constrain the optimization.
Due to the Markovian property of diffusion policies, the reverse KL can be estimated by 
\begin{align}
\KL(p_\theta|| p_{\text{ref}}) =    \mathbb E_{\rvx_{0:T} \sim p_\theta }\sum_{t} \KL (p_\theta(\cdot|\rvx_t,c) \| p_\text{ref}(\cdot|\rvx_t,c) ), \label{eq:kl_decomposed}
\end{align}
where $p(\cdot|\rvx_t,c)$ is the marginal distribution over $\rvx_{t-1}$. When both $p_\theta$ and $p_\text{ref}$ are diffusion models, the divergence can be analytically calculated by $\|\boldsymbol{\epsilon}_\theta(\rvx_t,t,c) - \boldsymbol{\epsilon}_\text{ref}(\rvx_t,t,c)  \|^2$. 
For the reward maximization term, classical approaches such as REINFORCE~\cite{sutton1999policy} points out that 
\begin{align*}
     \nabla_\theta  \mathbb E_{ p_\theta}[r(\rvx_0,c)] 
    &= \int_{\rvx_{0:T}} \nabla_\theta   p_\theta(\rvx_{0:T}|c)r(\rvx_0,c) \mathrm d\rvx \\
    &=  \mathbb E_{\rvx_{0:T} \sim p_\theta(\cdot|c)} [\nabla_\theta \log  p_\theta(\rvx_{0:T}|c)r(\rvx_0,c)].
\end{align*}
To optimize $\mathbb E[r(\rvx_0,c)]$, we can compute its gradient from the log probability of samples from $p_\theta$. 
In the context of diffusion, $\log p_\theta(\rvx_{0:T}|c)$ can be calculated by decomposing it into the sum of log probabilities at each step $t$.
Empirically, GRPO~\cite{shao2024deepseekmath} introduces relative advantages to stabilize the training. Specifically, for each condition $c$, the algorithm will sample a \textit{rollout group} of $\rvx_0^1,\cdots,\rvx_0^N$ from $p_\theta(\rvx_0|c)$ and replace $r(\rvx_0^n)$ by the advantage term $A(\rvx_0^n, c) \triangleq \frac{r(\rvx_0^n, c) - \mathrm{mean}} {\mathrm{std}}$, where $\mathrm{mean}, \mathrm{std}$ are within the group. While not theoretically grounded, advantages enable models to focus on good samples over bad samples rather than on absolute reward values, and are widely adopted.


\paragraph{Old policy.}Existing algorithms often maintain an ``old'' policy $p_{\theta_\text{old}}$ for sampling, and optimize the new policy $p_{\theta}$ with the reweighted reward $\frac{p_{\theta}(\rvx_0|c)}{p_{\theta_\text{old}}(\rvx_0|c)} r(\rvx_0,c)$. We still regard this as an on-policy approach because the sampling policy $p_{\theta_\text{old}}$ is periodically updated to $p_{\theta}$. As we show below, the underlying issue that causes reward hacking persists.




\section{Methodology}
\label{sec:framework}


\subsection{Motivation}
\label{subsec:motivation}
In generative modeling, reward maximization is not the purpose, but rather the means to learn $p_\theta$ that reweighs the original $p_\reft$ towards human preferences. 
Existing algorithms anchor $p_\theta$ to $p_\reft$ by either leveraging the reverse KL divergence (FlowGRPO~\cite{liu2025flow}) or by adopting heuristics such as a fixed initial random seed (DanceGRPO~\cite{xue2025dancegrpo}).
Unfortunately, they exhibit various types of hacking behaviors due to their intrinsically ineffective regularization. 


\begin{figure}[t]
    \centering
    \includegraphics[width=0.9\linewidth]{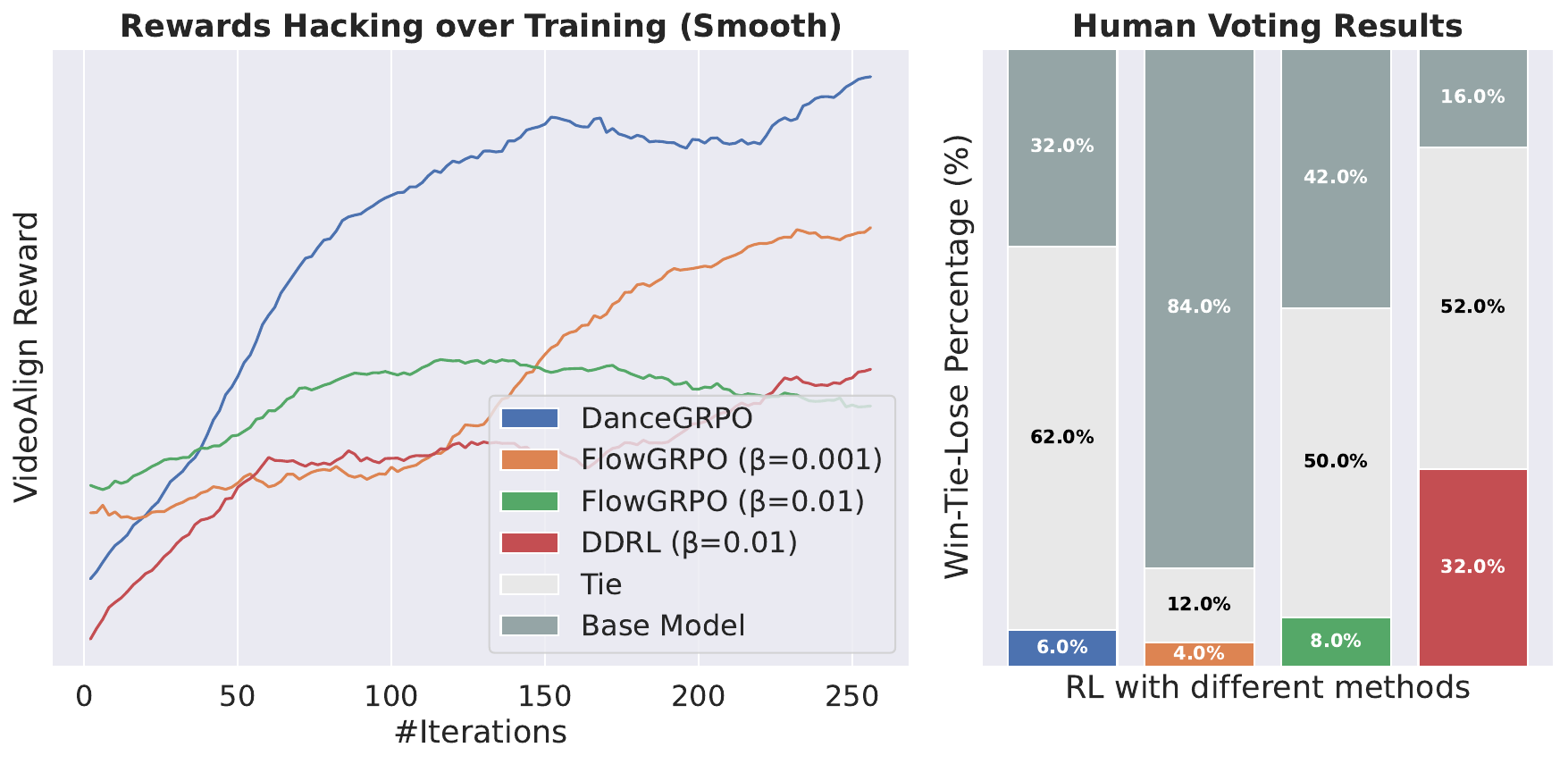}
    \vspace{-10pt}
    \caption{(Left) Reward hacking during training. All methods are trained from the same base model using the same dataset for 256 iterations. While DanceGRPO and FlowGRPO achieve higher rewards, (Right) humans consistently prefer videos generated by the base model. By contrast, \alg improves rewards and its generations are more likely preferred than the base model's generations.}
    \label{fig:hacking_and_vote}
    \vspace{-10pt}
\end{figure}

To illustrate this, we post-train Cosmos2.5~\cite{agarwal2025cosmos}, a physical world foundation model that aims to generate physically reasonable videos, using different methods. As shown in \cref{fig:video_hacking}, DanceGRPO optimizes rewards by generating over-colorized, over-stylized videos that are unrealistic, ignoring the prompt requirements. On the contrary, FlowGRPO generates blurry, jittering videos that are visually unacceptable. 
Notice that the phenomenon persists even when we carefully tune their hyperparameters such as regularization weight $\beta$, training iterations, or learning rate (\cref{sec:ddrl}).
As shown in \cref{fig:hacking_and_vote}, double-blind human voting between videos generated by the base model and those generated by post-trained models reveals that the original videos are much preferred, albeit that the post-trained models have higher rewards. 
A similar phenomenon is observed in image generation, where as shown in \cref{fig:t2i}, existing methods hack the text rendering rewards by generating over-simplified images that lose original realism.

\paragraph{Analysis on the reverse KL.} We are motivated to analyze why hacking happens even though we regularize with $\KL (p_\theta\| p_\reft)$. While an ideal reference model is sufficient to constrain the optimization and prevent weird samples in theory, practical models can be ``adversarially hacked'' due to the \textit{on-policy} KL calculation, where as shown in \cref{eq:kl_decomposed}, the log probability ratio is computed on $\rvx_t$ sampled from $p_\theta$. 
As the policy $p_\theta$ is driven by reward maximization, it gradually generates intermediate states $\rvx_t$ that are out-of-distribution for the reference model $p_{\text{ref}}$. In these unexplored regions of the latent space, the reference model may not have been well trained, making its regularization signal unreliable. The multi-step Markovian sampling can exacerbate the problem since the divergence at each step may be small while the accumulated shift of $\rvx_t$ is significant. 
\begin{figure}[t]
    \centering
    \includegraphics[width=\linewidth]{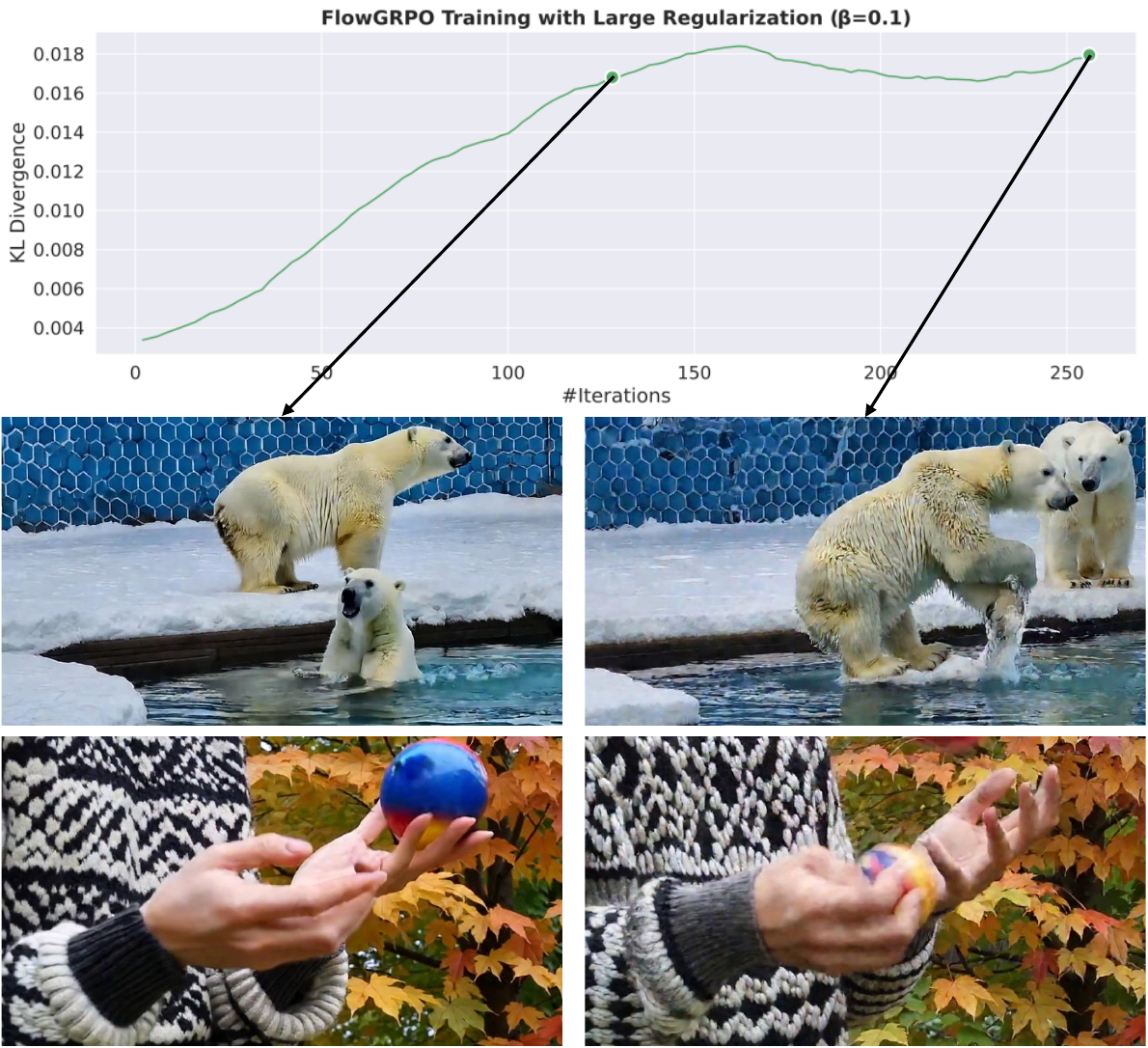}
    \vspace{-10pt}
    \caption{Even with significantly larger $\beta$ and the KL divergence remains stable throughout the training, unrealistic noise textures still appear in videos generated by the latter checkpoint.}
    \label{fig:largebeta}
    \vspace{-10pt}
\end{figure}

We empirically demonstrate our analysis by post-training with a significantly large coefficient $\beta = 0.1$. As shown in \cref{fig:largebeta}, while the KL divergence stabilizes since the middle of the training, the generated videos exhibit noise textures after longer training, implying that controlling the reverse KL divergence is insufficient to regularize the policy. Indeed, the on-policy sampling, imperfect reference, and non-verifiable rewards jointly make the existing regularization incapable of supporting effective post-training algorithms for diffusion models. 

\begin{algorithm*}[t]
\caption{Data-regularized Diffusion Reinforce Learning (\textbf{DDRL})}
\label{alg:train}
\textbf{Input:} Initial policy $\theta$, data sampler $\tilde p_\text{data}$, reward model $r$, training steps $\mathcal T \subset [0,\cdots, T-1]$, coefficient $\beta$, rollout size $N$.
\begin{algorithmic}[1]
\For {each training iteration}
\State  Sample $c$, sample $\tilde \rvx_0 \sim \tilde p_\text{data}(\cdot|c)$ 
\State Rollout $\{\rvx^n_t\}_{n=1}^N$ with $p_\theta$, get reward $r^n = r(\rvx_0^n, c)$, compute $A^n = \frac{r^n - \mathrm{mean}}{\beta \cdot \mathrm{std}}$.
\For {$t \in \mathcal T$}
\State Sample $\boldsymbol{\epsilon}^n$, get noisy $\tilde \rvx_t^n$ based on $(\tilde \rvx_0, \boldsymbol{\epsilon}^n, t)$.
\State Optimize $L^n_t = \|\boldsymbol{\epsilon}_\theta(\tilde \rvx_t^n, t, c) - \boldsymbol{\epsilon}^n \| - A^n \log p_\theta(\rvx_t^n|\rvx^n_{t+1}, c)  $ \algorithmiccomment{Simplied loss from $\cref{eq:ddrl_objective_diff}$}

\EndFor
\State Update $\theta$ via gradient descent w.r.t. all $L_t^n$.
\EndFor

\end{algorithmic}
\end{algorithm*}

\subsection{The \alg Framework}



Motivated by the problem of on-policy sampling in diffusion RL, a straightforward solution is to regularize with samples from the reference model (off-policy samples), i.e., to replace $\KL(p_\theta \| p_\reft)$ with the forward KL divergence $\KL (p_\reft \| p_\theta) $, which is computed using $\rvx_t$ sampled from $p_\reft$. This naive replacement will lead to an incorrect optimum different from \cref{eq:target_dist}, with more details in~\cite{gx2025kl}. To effectively leverage the off-policy regularization while maintaining the precise objective, this paper proposes a novel, principled, and robust forward KL divergence-based RL framework, named \textbf{Data-regularized Diffusion Reinforcement Learning} (\alg). Specifically, given a $p_\reft$, \alg maximizes $ {\mathcal J}_\alg(p_\theta)$ defined as:
\begin{equation}
\mathbb{E}_{p_\theta(\rvx_0 | c)}\Big[\lambda \Big( {\frac{r(\rvx_0,c) - Z}\beta}  \Big) \Big] - \KL(\tilde p_{\reft} || p_\theta),
    \label{eq:ddrl_objective}
\end{equation}
where $\lambda(x) = -\exp(-x)$ is a monotonic transform function, and constant $Z = \beta \log \mathbb E_{ p_\reft(\rvx_0|c)} [\exp\{r(\mathbf{x},c)/\beta \}]$ is an ``average reward'' that converts absolute rewards to relative advantages as inspired by GRPO. 
The core difference lies in $\tilde p_\reft$, which is defined as the \textit{forward distribution} specified in \cref{eq:add_noise}:
\begin{align*}
    \tilde p_\reft(\rvx_0| c) &= p_\reft(\rvx_0|c). \\
    \tilde p_\reft(\rvx_t | \rvx_{t-1}) &= \mathcal{N}(\sqrt{1-\beta_t} \rvx_{t-1}, \beta_t \mathbf{I}).
\end{align*}
Notably, \cref{eq:ddrl_objective} differs from \cref{eq:rl_objective} in two fundamental ways. First, it adopts the forward KL divergence, which enables regularization with off-policy samples from $p_\reft$; Second, it is based on the forward process $\tilde p_\reft$, which is different from the denoised process $p_\reft$.  
Below, we show that this principled formulation elegantly connects to diffusion training, post-train integration, and classical RL objectives.

\paragraph{Further transformation.} 
Using $\KL(\tilde p_\reft \| p_\theta) $ not only targets the hacking phenomenon brought by on-policy sampling, but also bridges to \cref{eq:diffusion_KL}, where $q$ is similarly defined from the data distribution.
This connection enables us to further transform $\KL(\tilde p_{\reft} || p_\theta)$ to the standard diffusion loss $\mathcal L(\theta; \tilde p_\reft(\rvx_0|c))$, similar to the conversion from \cref{eq:diffusion_KL} to \cref{eq:diffusion_loss}.
Thereafter, we only use the reference model for \textit{sampling}, instead of for computing intermediate variables as required in classical RL objective \cref{eq:rl_objective}, and it can flexibly be \textit{any} sampler.
Specifically, when the reference model is fine-tuned with a high-quality, accessible dataset $p_\text{data}$, users can directly compute the diffusion loss on this dataset, which is exactly how it is trained. 
In contrast, when the dataset used to trained the reference model is not available, users can generate synthetic data upfront and post-train with this data.
This leads to an equivalent objective $\tilde {\mathcal J}_\alg $, where we replace $\tilde p_\reft$ with $\tilde p_\text{data}$ to emphasize its nature as a sampler:
\begin{equation}
\mathbb{E}_{ p_\theta(\rvx_0 | c)}\Big[\lambda \Big( {\frac{r(\rvx_0,c) - Z}\beta}  \Big) \Big] - \mathcal L(\theta; \tilde p_\text{data}(\rvx_0|c)).
    \label{eq:ddrl_objective_diff}
\end{equation}



\begin{theorem}
\label{thm:optimal_policy}
Maximizing \cref{eq:ddrl_objective_diff} is equivalent to maximizing \cref{eq:ddrl_objective}, and its optimal policy $p_\theta^*(\mathbf{x})$ satisfies
\begin{equation}
    p_\theta^*(\rvx_0|c) \propto \tilde p_{\text{data}}(\rvx_0|c) \exp\left(r(\rvx_0,c)/\beta\right).
\end{equation}
\end{theorem}
The proof can be found in the appendix. \cref{thm:optimal_policy} demonstrates that \alg enables direct regularization with data while maintaining the correct optimal distribution (\cref{eq:rl_objective}). 
Notice that $\tilde p_\text{data}$ here can generally be a real data distribution or a synthetic distribution such as $p_\reft(\rvx_0|c)$, in which case the optimum goes back to $p_\reft(\rvx_0|c) \exp(r(\rvx_0,c) / \beta)$ as in traditional RL.
It enables direct utilization of real data in RL through diffusion loss, which is not possible in reverse KL-based approaches. 

\paragraph{\alg as a post-train integration.}Remarkably, \cref{eq:ddrl_objective_diff} points out the intrinsic nature of \alg as an elegant integration of supervised fine-tuning (via diffusion loss minimization) and reinforcement learning (via reward maximization).
Indeed, this combination has also been studied in recent LLM post-training works~\cite{chen2025beyond,lv2025towards}, offering cross-justification of over methodology. 
\cref{thm:optimal_policy} theoretically justifies for the first time the merit of combining two post-training stages, and its proof can be transferred to auto-regressive generative models as well. In short, \alg not only offers a theoretically precise formulation but also leads to an empirically robust and unified implementation. 

\subsection{Practical Implementation}
\label{subsec:implementation}

We present the pseudo-code of our practical implementation of \cref{eq:ddrl_objective_diff} in \cref{alg:train}.
For diffusion loss minimization, similar to standard diffusion training, we omit the coefficient $w_t$ for the sake of stability. For reward maximization, we omit the monotonic transformation $\lambda(\cdot)$ such that the training reward scales are consistent and comparable between \alg and baselines. We compute $Z$ using the average reward within the rollout group, and additionally divide the standard deviation. 
\paragraph{Timesteps \& Guidance.} We use $\mathcal T$ to denote the timesteps where \alg optimizes the objective. Similar to the observation in \cite{xue2025dancegrpo}, we find that optimizing half of the steps is sufficient, and \alg simply optimizes once every two steps. In terms of guidance, FlowGRPO utilizes classifier-free guidance (CFG) while DanceGRPO does not use CFG whenever possible. Following the diffusion training standard and recent studies~\cite{zheng2025diffusionnft}, \alg does not adopt CFG, and drops the condition $c$ when calculating the diffusion loss with $20\%$ probability.

\begin{table*}[t]
\centering
\caption{\textbf{DDRL aligns model with human preference better without severe reward hacking.} ``T2V'' and ``I2V'' stand for ``Text to Video'' and ``Image to Video''. $\Delta$-Vote denotes the difference in human preference from pairwise video comparisons, calculated as the win percentage of the corresponding method minus the win percentage of \alg. We \red{mark a reward} if its generated videos are less preferred by humans but the reward value is higher in this setting, which generally indicates reward hacking.}
\label{tab:main_table}
\vspace{-10pt}
\resizebox{\textwidth}{!}{%
\begin{tabular}{llcccccccc}
\toprule
\multirow{2}{*}{\textbf{Base Model}} & \multirow{2}{*}{\textbf{Method}} & \multicolumn{2}{c}{VideoAlign T2V} & \multicolumn{2}{c}{VideoAlign I2V} & \multicolumn{2}{c}{VBench T2V} & \multicolumn{2}{c}{VBench I2V} \\ \cmidrule(lr){3-4} \cmidrule(lr){5-6} \cmidrule(lr){7-8} \cmidrule(lr){9-10}
 & & $\Delta$-\textbf{Vote}$\uparrow$ (\%) & \textbf{Reward} & $\Delta$-\textbf{Vote}$\uparrow$ (\%) & \textbf{Reward} & $\Delta$-\textbf{Vote}$\uparrow$ (\%) & \textbf{Reward} & $\Delta$-\textbf{Vote}$\uparrow$ (\%) & \textbf{Reward} \\ \midrule
\multirow{4}{*}{Cosmos2.5-2B} & 
w/o RL       & -22.9 & 0.408 & -4.2 & 0.079 & -8.9 & 0.830 & -15.7 & 0.810 \\
 & DanceGRPO & -10.5 & \red{0.715} & -6.4 & \red{0.254} & -6.9 & \red{0.849} & -8.0 & \red{0.822} \\
 & FlowGRPO  & -6.7 & 0.408 & -13.0 & 0.069 & -4.2 & 0.830 & -9.6 & 0.807 \\
 & DDRL      & 0 & 0.604 & 0 & 0.177 & 0 & 0.842 & 0 & 0.819 \\ \midrule
\multirow{4}{*}{Cosmos2.5-14B} & 
w/o RL       & -8.4 & 0.359 & -12.5 & 0.058 & -4.5 & 0.820 & -4.5 & 0.813 \\
 & DanceGRPO & -4.9 & 0.494 & -5.9 & \red{0.160} & -3.1 & 0.834 & -4.2 & 0.818 \\
 & FlowGRPO  & -6.7 & 0.476 & -11.6 & 0.128 & -7.3 & 0.829 & -5.4 & 0.808 \\
 & DDRL      & 0 & 0.555 & 0 & 0.134 & 0 & 0.835 & 0 & 0.819 \\ \bottomrule
\end{tabular}
}
\vspace{-10pt}
\end{table*}

\paragraph{Memory \& Efficiency.} Both real and synthetic data can be collected upfront, so \alg needs neither a reference model nor an old model during training, thus saving memory by a large margin. That said, all importance sampling-based reward maximization alternatives are naturally compatible with \alg, as the core improvement lies in adopting the diffusion loss.
In terms of efficiency, the vanilla version of \cref{alg:train} optimizes the diffusion loss of each single data point $N|\mathcal T|$ times ($N$ rollouts, $|\mathcal T|$ steps). We show in \cref{tab:ablation} that this can be substantially reduced to $N$ times by sampling $t$ independently and optimizing the diffusion loss only once.
As a result, the number of network evaluations (NFEs) per data of \alg is similar to DanceGRPO (w/o KL divergence), and a half compared to FlowGRPO.



\begin{table}[!b]
\vspace{-10pt}
    \centering
    \caption{Hyperparameters used in different methods.}
    \label{tab:hyper}
    \vspace{-10pt}
    \begin{tabular}{lcccc}
    \hline

       \textbf{Method} & $T$ & $|\mathcal T|$ & $\beta$ & use $p_\reft$? \\
\hline
       DanceGRPO  &  20 & 12 & $0$ & $\times$ \\
       FlowGRPO   &  10 & 10 & $[0.001, 0.1]$ & \checkmark \\
       \alg & 20 & 10 & $0.01$ & $\times$\\
\hline
    \end{tabular}
\end{table}

\section{Experiments}
\label{sec:experiments}

\subsection{Setting}

\paragraph{Models.} For video generation, we use Cosmos2.5~\cite{ali2025world,agarwal2025cosmos} 2B and 14B as the base models due to their capability in generating physically reasonable scenarios. 
We use two reward models: VideoAlign~\cite{liu2025improving} finetunes Qwen2-vl~\cite{wang2024qwen2} to evaluate the visual quality, motion quality, and text alignment of a video, and we use the average of three scores as our reward. VBench~\cite{huang2024vbench} contains a wide range of model-based metrics. We use the average of subject consistency, background consistency, motion smoothness, aesthetic quality, and imaging quality (five scores).

\paragraph{Training.} We train with 256 H100 GPUs for Cosmos2.5-2B and 1024 GPUs for Cosmos2.5-14B, with rollout size $N=8$ and batch size 16. We host all reward models in a remote server with 512 GPUs to calculate rewards asynchronously. 
Throughout the paper, we use the same 1:1 mixture of Text to Video (T2V) and Image to Video (I2V) data for rollout.
We implement baselines based on original papers and codebases, with hyperparameters specified in \cref{tab:hyper}. Notice that a larger $|\mathcal T|$ naturally results in better optimization. For all methods, we accumulate the gradient over all $t \in \mathcal T$ and conduct one full parameter update at the end of each iteration.
We consume approximately one million GPU hours. More details are presented in the appendix.

\paragraph{Evaluation.} All reported measurements are tested on a PAI-Bench~\cite{PAIBench2025}, a high-quality test dataset containing five categories (human, industry, robotics, driving, others) and 1044 prompts. For all methods, we generate videos using the trained models under the T2V and I2V (same conditional images) settings with classifier-free guidance. 
Since we aim to analyze the hacking phenomenon, comparing only the reward values becomes misleading, and we conduct large-scale double-blind \textbf{human voting} across fifteen researchers.
Specifically, under each setting, we compare \alg with FlowGRPO, DanceGRPO, and the base model separately by asking voters to select which generated videos are better (or tie) via a side-by-side comparison website. 
Voters are asked to vote based on their preference to the generated videos.
We report the difference in the preference percentages (denoted as $\Delta$-\textbf{Vote}): if \alg wins on 40.0\% videos and loses on 18.0\%, then the reported $\Delta$-Vote is 32.0. 
In total, we gather $\sim$9600 comparison data as the ground-truth to help reflect the quality of a method~\cite{skalse2022defining}: we expect a good method to increase rewards and improve/maintain good human preference at the same time.

\subsection{Superiority of \alg}
\label{sec:ddrl}

Over 15\% of the computation is spent on selecting the best hyperparameters for baseline methods to ensure that large-scale human voting is informative. This includes the number of training iterations and the coefficient $\beta$. As we demonstrate in \cref{fig:hacking_and_vote,fig:largebeta}, both DanceGRPO and FlowGRPO (regardless of $\beta$ ranging across two magnitudes) hack severely and become worse than the base model after training for 256 iterations. To reduce the phenomenon, we train all methods only for 128 iterations and use the exponential moving average (EMA) model weights to generate test videos. 
In addition, we also carefully sweep the regularization parameter $\beta \in [0.001, 0.1]$ for FlowGRPO, and select $\beta = 0.01$, consistent with their paper.
While this value does not prevent reward hacking, it mitigates the most severe artifacts, despite stifling legitimate reward growth. 
For \alg, we simply use the same $\beta$ as well.
Lastly, for 14B post-training, this is insufficient to prevent hacking, so we further reduce the learning rate to 3e-6 for all methods. 

We present a summary of voting results and rewards under two reward models, two generation settings, and two base models in \cref{tab:main_table}, with details provided in the appendix.
Simultaneously across all settings, \alg not only improves rewards over the base model, but is also more preferred than any other baselines or the base model, with $\Delta$-Vote always being negative indicating that \alg-generated videos are consistently more favorable. Importantly, \alg is the only method that improves rewards in a way that voters prefer, whereas DanceGRPO often hacks the reward models by generating unpreferred videos, which nevertheless receive higher rewards. 
In addition, the advantage in 14B experiments is smaller due to the use of a smaller learning rate to mitigate baselines' hacking behaviors, and the performance gap will enlarge as we continue post-training.

\begin{figure}[t]
    \centering
    \includegraphics[width=\linewidth]{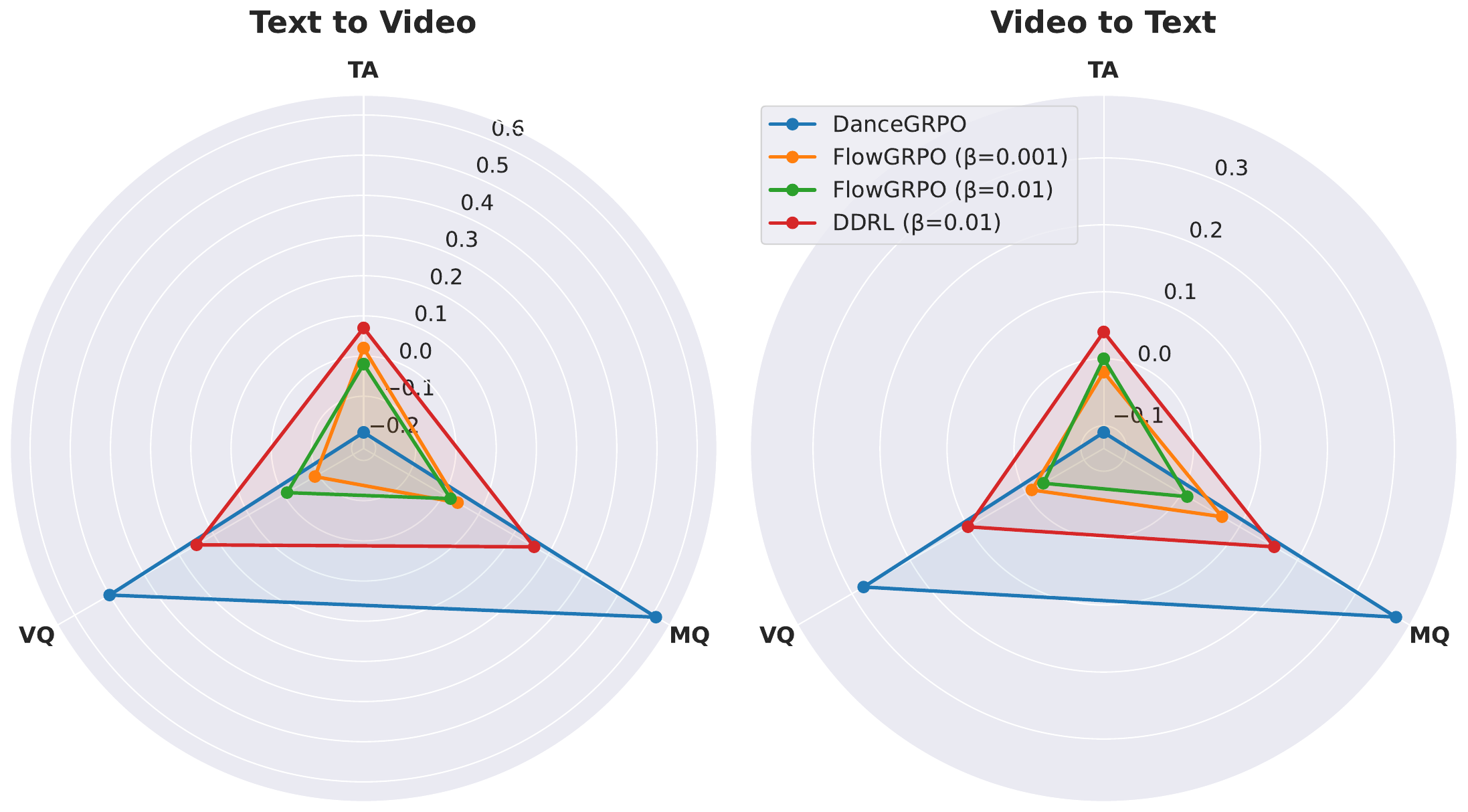}
    \caption{Breakdown of the the increase/decline of text alignment (TA), visual quality (VQ), and motion quality (MQ), after post-training the 2B model with different methods. DanceGRPO shows a 16\% (T2V) and 28\% (I2V) decrease in the TA score, despite its average reward is the highest. This aligns with the human voting result where its generated videos are not preferred. \alg is the only algorithm that brings ``Pareto improvement''.
    }
    \label{fig:hack_radar}
    \vspace{-10pt}
\end{figure}

\paragraph{Reward hacking analysis.}
We carefully analyze why existing methods do not work well by presenting a breakdown of per-score-increase after RL in \cref{fig:hack_radar}.
DanceGRPO achieves its high average reward at the cost of sacrificing one score, with text alignment increase being negative after post-training.
This trade-off is a signal of reward hacking, indicating that one property is being compromised to exploit the reward.
\alg, by regularizing directly with the data-anchored diffusion loss, achieves a Pareto improvement across all scores. We believe that the ability to enhance alignment without sacrificing other generative properties is a highly appealing and robust feature in RL.

Notably, our large-scale human voting is labor-intensive and not always be adoptable. A reliable way to automatically detect reward hacking is highly beneficial. Although we have yet to figure it out, several metrics could be indicative. For instance, if the diffusion loss of a post-trained checkpoint increases by more than 10\%, its outputs are mostly hacked; a steep increase in rewards or a sharp decrease in standard deviation are also potential signals.

\begin{table}[ht]
\centering
\caption{Ablations on training iteration and on the scheduler of $t$ for diffusion loss, using VideoAlign as the reward model. \alg optimizes the diffusion loss for each timestep $t \in \mathcal T$, resulting in an additional of $N |\mathcal T|$ network evaluations. ``Reduced'' only compute the loss once by randomly sampling $t$.}
\label{tab:ablation}

\vspace{-8pt}
\small
\begin{tabular}{lcccc}
\hline
\textbf{$\mathcal T$} \textbf{schedule} & \textbf{Iteration} & \textbf{NFEs per $c$} & \textbf{T2V} & \textbf{I2V} \\
\hline
\alg & 128 & $2N|\mathcal T|$ & 0.604 & 0.177  \\
\alg & 256 & $2N|\mathcal T|$ & 0.627 & 0.180  \\
Reduced & 256 & $N(1+|\mathcal T|)$ & 0.636 & 0.143 \\
\hline
\end{tabular}
\vspace{-15pt}
\end{table}

\paragraph{Ablations on \alg.}
Our ablation studies in \cref{tab:ablation} reveal two key findings. First, while our main results use a conservative iteration, \alg can continuously improve rewards without hacking even when training iterations are doubled. Additional human voting confirms that this extended training does not degrade user preference. Second, computing the diffusion loss only once per step (by randomly sampling $t$) is as effective as optimizing the loss at every timestep in the set $\mathcal T$. This makes \alg the first algorithm to adopt explicit regularization while matching the high computational efficiency of regularization-free methods.

\subsection{\alg as post-training integration}
\label{sec:exp_unify}

\begin{figure}[t]
    \centering
        \includegraphics[width=\linewidth]{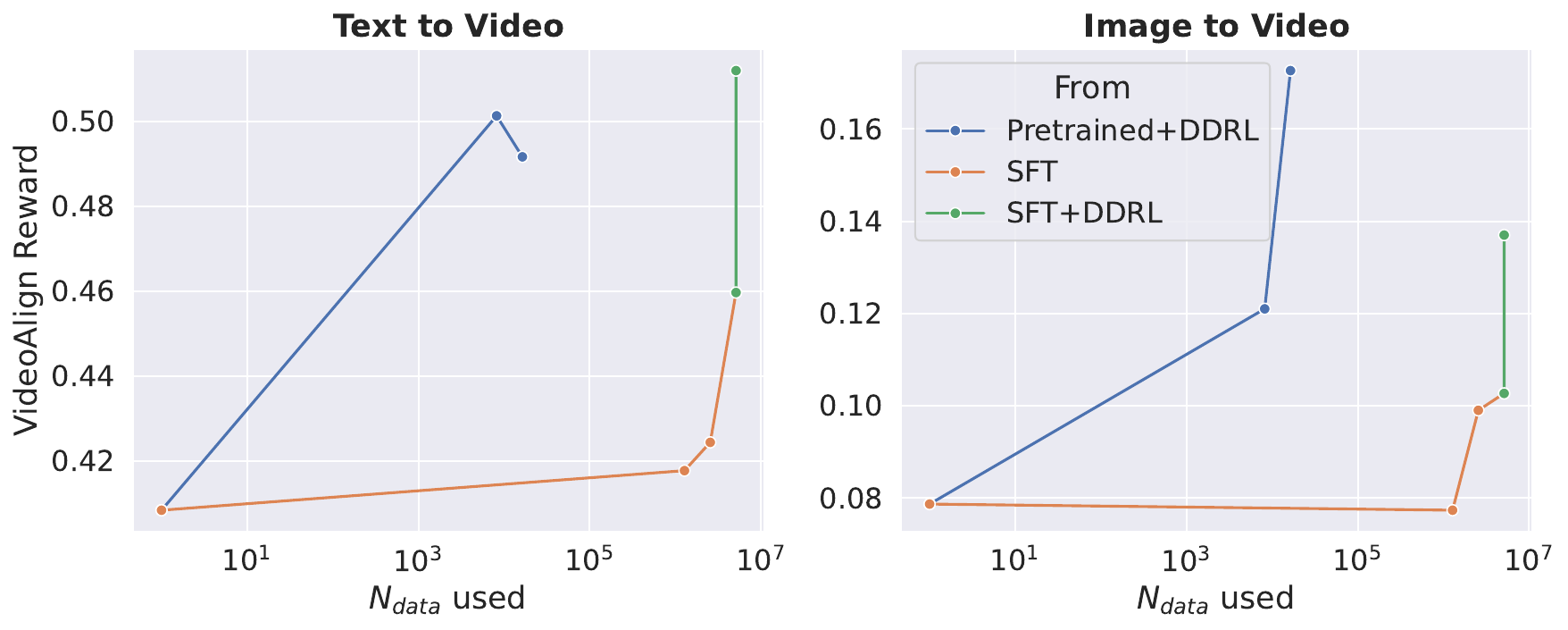}
    \caption{VideoAlign reward after post-training with \alg from (1) pretrained model and (2) SFT model. The former achieves comparably high rewards with significantly higher data efficiency. }
    \label{fig:base_vs_sft}
    \vspace{-10pt}
\end{figure}

We have pointed out that \alg essentially combines RL with SFT as its objective.
This raises a natural question: can it be used to directly post-train a model, replacing the standard SFT-then-RL pipeline? We compare two training strategies: (1) a conventional approach where we SFT a vanilla pretrained model on a high-quality dataset for 20K iterations, then post-train with \alg; (2) we apply \alg directly to the vanilla pretrained checkpoint using the same dataset. Interestingly, as shown in \cref{fig:base_vs_sft}, the direct \alg approach achieves comparably high rewards and low diffusion loss (0.119 v.s. 0.121) to the SFT-then-RL pipeline. While the direct approach saves the entire 20K iteration SFT step, we note that the on-policy RL rollouts—which reuse data multiple times—remain the dominant computational cost in both settings. Nevertheless, this result demonstrates a strong potential to simplify and unify the diffusion post-training paradigm, and we encourage future research to examine this possibility more closely.

\subsection{\alg with synthetic data}

We have demonstrated the superiority of \alg with data regularization, and below we show that \alg works effectively in the absence of real data.
Specifically, we apply \alg to text-to-image (T2I) tasks based on SD3.5-Medium~\citep{esser2024scaling}, with settings aligned with FlowGRPO. 
We focus on improving the OCR reward for visual text rendering, which is vulnerable to reward hacking. We finetune with LoRA ($\alpha=64$, $r=32$) and use synthetic data sampled from the base model for diffusion loss regularization.

\begin{table}[ht]
\centering
\caption{Human voting results, OCR reward, and three OOD rewards after post-training SD3.5-Medium with different methods. We similarly \red{mark the OCR reward} if it is achieved by sacrificing human preference, e.g., generations are over-simplified or cartoonish compared with the base model.}
\label{tab:t2i}
\vspace{-10pt}
\resizebox{\linewidth}{!}{
\begin{tabular}{lcc|ccc}
\hline \textbf{Method} & $\Delta$-\textbf{Vote}$\uparrow$ (\%) & \textbf{OCR}$\uparrow$ & \textbf{ClipScore}$\uparrow$ & \textbf{PickScore}$\uparrow$ & \textbf{ImageReward}$\uparrow$ \\
\hline
w/o RL & - & 0.566 & 0.321 & 0.865 & 1.13  \\
DanceGRPO & -53.0 & \red{0.846} & 0.309 & 0.837 & 0.80  \\
FlowGRPO & -22.1 & \red{0.845} & 0.313 & 0.855 & 1.03 \\
DDRL & 0 & 0.823 & 0.320 & 0.865 & 1.14 \\
\hline
\end{tabular}
}
\end{table}

\begin{figure}[!t]
    \centering
    \includegraphics[width=\linewidth]{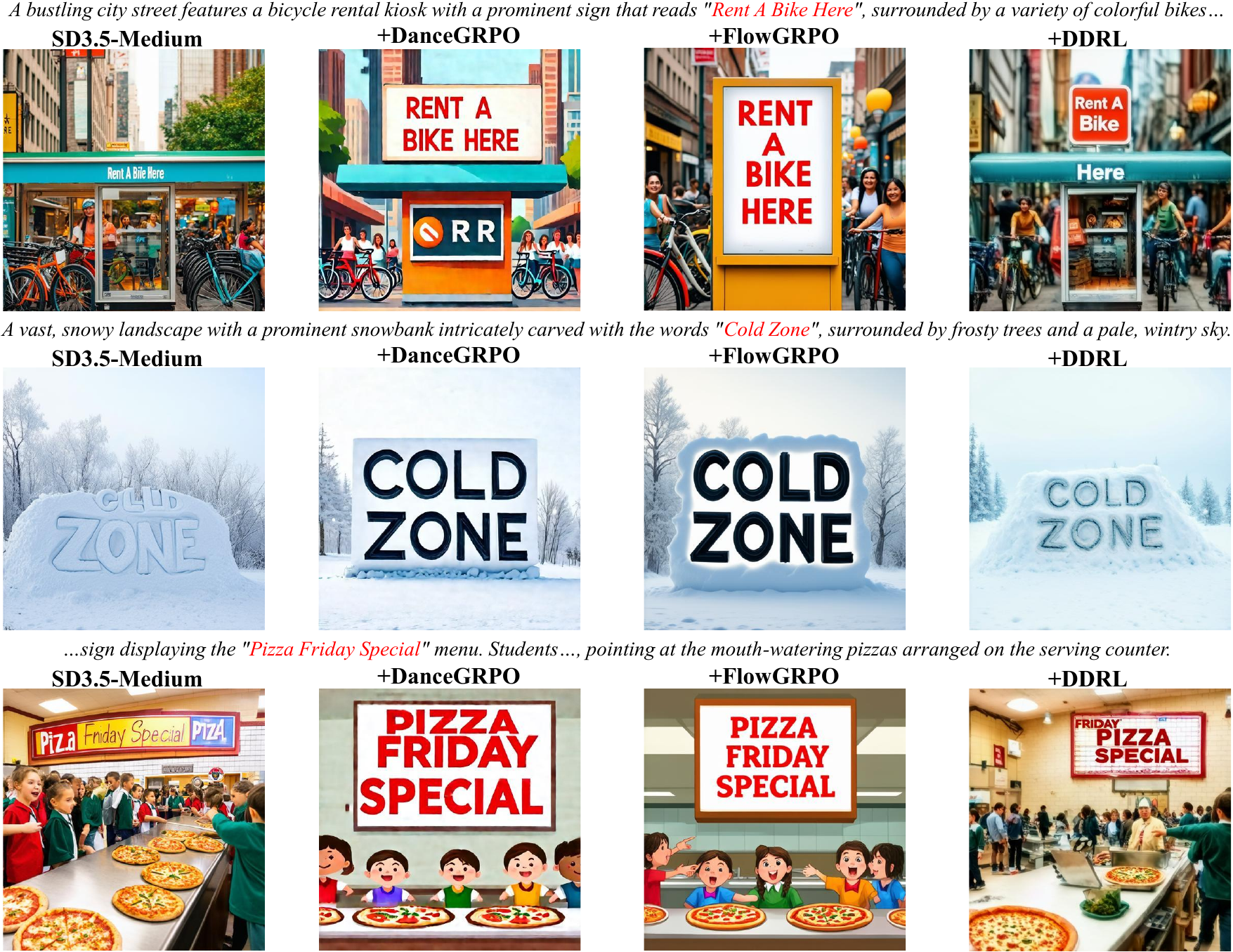}
    \caption{Generated images after RL on the OCR reward. With similarly high OCR accuracy, \alg additionally maintains fidelity to the base model in terms of style and realism. }
    \label{fig:t2i}
    \vspace{-15pt}
\end{figure}

As shown in Figure~\ref{fig:t2i} and Table~\ref{tab:t2i}, \alg exhibits both improved OCR accuracy and over 20\% human preference due to its fidelity to the base model's style and realism, even when we purely use synthetic data for regularization. In contrast, baseline methods tend to simplify images by using large, centralized captions.
We also demonstrate that \alg achieves high OOD rewards~\cite{hessel2021clipscore,kirstain2023pick,xu2023imagereward}, which shows its robustness in preserving generation diversity.


\section{Related Works \& Conclusions}
\paragraph{RL and Guidance.} Both RL and guidance align generative models with human preferences and sample from $p_\reft(\rvx_0 |c) \exp(r(\rvx_0, c) / \beta)$. Classical classifer-guidance~\cite{dhariwal2021diffusion} trains a time-dependent classifier $r_t(\rvx_t,c)$ that evaluates the score of noisy variable $\rvx_t$, and trains the diffusion models with $r_t$. Later, classifier-free guidance~\cite{ho2022classifier} encodes condition $c$ during training and injects guidance by adding $\boldsymbol{\epsilon}(\rvx_t,t,c) - \boldsymbol{\epsilon}(\rvx_t,t,\emptyset)$ during inference. This effective and flexible framework is a \textit{de facto} standard. Recently, training-free guidance~\cite{ye2024tfg} is proposed to conduct controllable generation using an off-the-shelf reward model $r(\rvx_0,c)$ without additional training. Most approaches predict outputs during the denoising process and guide the generation with the gradient of the reward model. RL stands for the opposite: it requires additional training but does not require differentiable rewards. A deeper connection between RL and guidance remains to be explored.

\paragraph{Reward hacking.} Reward hacking, or reward gaming, is a long-lasting issue that goes far beyond the RL domain~\cite{skalse2022defining}. It is closely related to Goodhart's Law proposed by the economist Charles Goodhart~\cite{chrystal2003goodhart}, who states that a measure ceases to be a good measure when it becomes the target. This is exactly what happens with reward models in RL. While the law has also been studied in AI domains~\cite{clark2016faulty,karwowski2023goodhart}, reward hacking remains a primary problem that hinders the development~\cite{amodei2016concrete}. A more principled metric to detect reward hacking will be beneficial. 

\paragraph{Integrating SFT and RL.} Generative models are trained through a multi-stage pipeline, where SFT uses a high-quality dataset to tune the model and RL optimizes the policy for a learned reward that captures human preferences. \alg has pointed out the possibility of integrating the two post-training stages for diffusion models, and similar attempts have also been made for LLMs~\cite{chen2025beyond,lv2025towards}, although these are empirical and preliminary. Our study provides theoretical justification for this integration, showing that it could be beneficial to model robustness and data efficiency. 
Future research could propose an integrated post-training algorithm for LLMs based on the presented observations.

\paragraph{Conclusion.} Reward hacking in diffusion RL is an intrinsic issue of on-policy regularization. This paper proposes \alg, a novel framework with a theoretically robust objective and an empirically simple, stable, and highly scalable algorithm that combines on-policy reward maximization with diffusion loss minimization. Our extensive experiments demonstrate that \alg improves rewards and aligns with human preference by effectively alleviating reward hacking, establishing a principled and effective foundation for the future of post-training diffusion models.
\newpage
{
    \small
    \bibliographystyle{ieeenat_fullname}
    \bibliography{main}
}



\appendix
\clearpage
\setcounter{page}{1}
\newpage
\onecolumn
    \begin{center}
        \Large
        \text{\thetitle}\\
        \vspace{0.5em}Supplementary Material \\
        \vspace{1.0em}
        
    \end{center}

\section{Proof of \cref{thm:optimal_policy}}


\paragraph{Part 1: Optimal solution.}

We derive the optimal policy for the theoretical \alg\ objective defined in \cref{eq:ddrl_objective}.

Expanding the KL divergence term:
\begin{align}
    D_{\mathrm{KL}}(\tilde{p}_{\mathrm{ref}} || p_{\theta}) &= \int \tilde{p}_{\mathrm{ref}}(\rvx_{0:T}|c) \log \frac{\tilde{p}_{\mathrm{ref}}(\rvx_{0:T}|c)}{p_{\theta}(\rvx_{0:T}|c)} d\rvx_{0:T} \\
    &= \mathbb{E}_{\tilde{p}_{\mathrm{ref}}}[\log \tilde{p}_{\mathrm{ref}}(\rvx_{0:T}|c)] - \int \tilde{p}_{\mathrm{ref}}(\rvx_{0:T}|c) \log p_{\theta}(\rvx_{0:T}|c) d\rvx_{0:T}.
\end{align}

Maximizing $\mathcal{J}_\alg(p_{\theta})$ is equivalent to maximizing the simplified functional $\hat{\mathcal{J}}_\alg(p_{\theta})$

\begin{equation}
    \hat{\mathcal{J}}_{\alg}(p) = \int p_{\theta}(\rvx_{0:T}|c) \lambda\left(\frac{r(\rvx_0, c)-Z}{\beta}\right) d\rvx_{0:T} + \int \tilde{p}_{\mathrm{ref}}(\rvx_{0:T}|c) \log p_{\theta}(\rvx_{0:T}|c) d\rvx_{0:T}.
\end{equation}

The first term is linear in $p_{\theta}$, and the second term involving $\log p_{\theta}$ is strictly concave. Thus, the functional $\hat{\mathcal{J}}_{\alg}(p)$ is strictly concave with respect to the trajectory distribution $p_{\theta}(\rvx_{0:T}|c)$. This ensures that the stationary point derived below is the unique global maximizer.

We formulate the Lagrangian with the normalization constraint $\int p_{\theta}(\rvx_{0:T}|c) d\rvx_{0:T} = 1$:
\begin{equation}
    \mathcal{L}(p_{\theta}, \mu) = \hat{\mathcal{J}}_{\alg}(p_{\theta}) - \mu \left( \int p_{\theta}(\rvx_{0:T}|c) d\rvx_{0:T} - 1 \right).
\end{equation}
Taking the functional derivative w.r.t. $p_{\theta}(\rvx_{0:T}|c)$ and setting it to zero:
\begin{equation}
    \lambda\left(\frac{r(\rvx_0, c)-Z}{\beta}\right) + \frac{\tilde{p}_{\mathrm{ref}}(\rvx_{0:T}|c)}{p_{\theta}(\rvx_{0:T}|c)} - \mu = 0.
\end{equation}
Solving for $p_{\theta}(\rvx_{0:T}|c)$:
\begin{equation}
    p_{\theta}(\rvx_{0:T}|c) = \frac{\tilde{p}_{\mathrm{ref}}(\rvx_{0:T}|c)}{\mu - \lambda\left(\frac{r(\rvx_0, c)-Z}{\beta}\right)}.
\end{equation}
Substituting $\lambda(u) = -\exp(-u)$ and setting $\mu=0$ (which satisfies the constraint):
\begin{equation}
    p_{\theta}^*(\rvx_{0:T}|c) = \tilde{p}_{\mathrm{ref}}(\rvx_{0:T}|c) \exp\left(\frac{r(\rvx_0, c)-Z}{\beta}\right).
\end{equation}

To find the optimal sample distribution $p_{\theta}^*(\rvx_0|c)$, we marginalize out the intermediate steps $\rvx_{1:T}$. Recall that $\tilde{p}_{\mathrm{ref}}(\rvx_{0:T}|c)$ is defined by the forward process starting from the data distribution $\tilde p_{\mathrm{data}}$. Integrating over the transition probabilities $\rvx_{1:T}$ yields the desired result:

\begin{equation}
    p_{\theta}^*(\rvx_0|c) = \tilde{p}_{\mathrm{data}}(\rvx_0|c) \exp\left(\frac{r(\rvx_0, c)-Z}{\beta}\right) \propto \tilde{p}_{\mathrm{data}}(\rvx_0|c) \exp\left(\frac{r(\rvx_0, c)}{\beta}\right).
\end{equation}


\paragraph{Part 2: Equivalence of loss.}

We now prove that minimizing the regularization term $D_{\mathrm{KL}}(\tilde{p}_{\mathrm{ref}} || p_{\theta})$ in \cref{eq:ddrl_objective} is equivalent to minimizing the diffusion loss $\mathcal{L}(\theta; \tilde{p}_{\mathrm{data}})$ in \cref{eq:ddrl_objective_diff}. 

The forward KL divergence is:
\begin{equation}
    D_{\mathrm{KL}}(\tilde{p}_{\mathrm{ref}}(\cdot|c) || p_{\theta}(\cdot|c)) = \mathbb{E}_{\tilde{p}_{\mathrm{ref}}} \left[ \log \frac{\tilde{p}_{\mathrm{ref}}(\rvx_{0:T}|c)}{p_{\theta}(\rvx_{0:T}|c)} \right]
\end{equation}

Start from
\[
D_{\mathrm{KL}}\!\big(\tilde p_{\mathrm{ref}}\|p_\theta\big)
=
\mathbb E_{\tilde p_{\mathrm{ref}}}\!\left[
\log \tilde p_{\mathrm{data}}(\rvx_0)
+\sum_{t=1}^{T}\log q(\rvx_t\mid \rvx_{t-1})
-\log p(\rvx_T)
-\sum_{t=1}^{T}\log p_\theta(\rvx_{t-1}\mid \rvx_t)
\right].
\]

By the telescoping identity, we have 

\begin{equation}
\label{eq:telescope}
\sum_{t=1}^{T}\log q(\rvx_t\mid \rvx_{t-1})
=
\log q(\rvx_T\mid \rvx_0)
+
\sum_{t=2}^{T}\log q(\rvx_{t-1}\mid \rvx_t,\rvx_0),
\end{equation}

Hence we have
\[
D_{\mathrm{KL}}\!\big(\tilde p_{\mathrm{ref}}\|p_\theta\big)
=
\mathbb E\Big[
\underbrace{\log q(\rvx_T\mid \rvx_0)-\log p(\rvx_T)}_{D_{\mathrm{KL}}(q(\rvx_T\mid \rvx_0)\,\|\,p(\rvx_T))\ }
+
\sum_{t=2}^{T}\!\big(\log q(\rvx_{t-1}\mid \rvx_t,\rvx_0)-\log p_\theta(\rvx_{t-1}\mid \rvx_t)\big)
-\log p_\theta(\rvx_0\mid \rvx_1)
\Big]
+ C_0,
\]
with $C_0=\mathbb E[\log \tilde p_{\mathrm{data}}(\rvx_0)]$ independent of $\theta$.
Under the Gaussian assumptions and fixed $\tilde\beta_t$, each $t\ge2$ term is a KL between Gaussians with the same covariance, hence a quadratic in the difference of means; writing both means via $\varepsilon$ gives a weighted $\varepsilon$-MSE contribution with weight $w_t$.
The $t=1$ term equals
\(-\mathbb E\log p_\theta(\rvx_0\mid \rvx_1)\),
which is another quadratic that yields the weight $w_1$.
Collecting all terms produces \cref{eq:ddrl_objective_diff} with a constant $C$ that does not depend on $\theta$.
The optimizer equivalence follows since subtracting $C$ does not change $\arg\max_\theta$.

\section{Experiment Details}

In this section, we provide a detailed description of our experimental setup.

\subsection{Video post-training}
\paragraph{Model.}Except for the SFT-RL integration experiments that we will cover later, our video experiments are initialized from the NVIDIA Cosmos foundation model that is first pre-trained on a large-scale pre-training dataset, fine-tuned separately on six different categories of high-quality datasets, and eventually merged into a single checkpoint ready for RL training. Notice that this model is different from the Cosmos2.5 models released in \cite{ali2025world}, which is the checkpoint after they post-train the merged checkpoint with \alg, i.e., the same process we conduct in \cref{tab:main_table}. Both 2B and 14B models are dense transformer-based diffusion models, performed under the latent tokenized space that is encoded and decoded using the Wan-2.1 variational autoencoder~\cite{wan2025wan}. Both models are trained and evaluated at high resolution, where the output videos have resolutions within the ranges (1280, 720) and (720, 1280). All videos have 93 frames, which corresponds to 24 latent frames with a FPS of 16.
For the text encoder, we leverage Cosmos-Reason1~\cite{azzolini2025cosmos} and encode the input prompt $c$ into embedding vectors. This process follows exactly the pre-training stage, with more details provided in \cite{ali2025world}. 

\paragraph{Data.}All algorithms sample prompts (and videos for \alg) from the high-quality dataset that is used to SFT the model. This dataset includes videos of different categories, ranging from human dynamics, physics, robotics, autonomous driving, and smart spaces. The training dataset is \textit{i.i.d.} to the evaluation PAI-Bench dataset. We always post-train the model for within a thousand iterations, consuming no more than 8,000 unique data points, which is only a small fraction of the whole dataset.
This dataset is also used in the SFT-RL integration study to fine-tune the pre-trained model. 

\paragraph{Training recipe.}We distribute our model using FSDP2. To control the per-GPU memory usage, similar to the Cosmos pre-training strategy, we employ context parallelism (CP), where one data copy will be distributed to two GPUs for the 2B training and eight GPUs for the 14B training. All parameters used for RL are presented in \cref{tab:hyper}. During each iteration, we first synchronize data across GPU workers within the same rollout group, and then perform rollout in parallel. The output latents are sent to the reward server for reward calculation, which will be clarified in \cref{app:reward_server}. 
Each worker immediately starts computing the diffusion loss when rewards are computed asynchronously, thereby avoiding reward calculation from blocking the training process. Rewards are fetched until we compute the final loss for gradient calculation.

\begin{table}[]
    \centering
    \caption{Hyperparameters used for video post-training.}
    \label{tab:hyper}
    \begin{tabular}{lcc}
    \hline
    Model size     & 2B & 14B \\
    \hline
    Batch size (Number of unique condition $c$) & 16 & 16 \\
    Rollout size $N$ & 8 & 8 \\
    Number of GPUs per data instance & 2 & 8 \\
    Total GPUs used & 256 & 1024 \\
    Optimizer & \multicolumn{2}{l}{AdamW ($\beta_1 = 0.9, \beta_2 = 0.99$)} \\
    Learning rate & 1e-5 & 3e-6 \\
    
    \hline

    \end{tabular}
\end{table}


\paragraph{Evaluation details.} As we mentioned above, we adopt PAI-Bench~\cite{PAIBench2025} as our test dataset, with 1044 high-quality test samples (and initial conditioning frames) ranging across five categories (human, industry, robotics, driving, others). For each checkpoint after post-training with each method, we generate videos using 35 denoising steps and a guidance strength of 7 (for 2B) and 3 (for 14B), separately for the text-to-video and image-to-video settings. Once we have all videos generated, we evaluate their reward score using the reward server same as in the training stage.

For human voting comparison between \alg and each of the baseline methods (and w/o RL setting), we randomly sample a condition $c$ from the test dataset, and then present the videos generated under condition $c$ side-by-side (random order). The voter is asked to choose which video is preferred (or tie), and voters do not know their voting results until they finish voting. In total, 15 people participated in the voting, where half of the voters are authors of this paper and the other half did not know this paper at all.

\paragraph{SFT-RL integration details.} Our study of integrating SFT and RL in \cref{sec:exp_unify} shares the identical setting of our main experiments except for the initialization. Specifically, the first setting is initialized with the pre-trained checkpoint (no fine-tuning or model merging). For the second setting, we simplify the Cosmos2.5 training and only conduct a single fine-tuning on the post-training dataset (so results are different from those in \cref{tab:main_table}). We fine-tuned the pre-trained model for 20,000 iterations using 512 GPUs. We optimize with AdamW ($\beta_1 = 0.9,\beta_2 = 0.999$)  with a learning rate of 3e-5.

\subsection{Video reward server}
\label{app:reward_server}

To ensure scalability, flexibility, and computational efficiency, we offload reward computation to a dedicated external service, decoupling it from the core generative pipeline. This service supports diverse video quality and alignment metrics—including VBench, VideoAlign, and Cosmos-Reason1-7B-Reward—enabling modular, extensible reward modeling. To minimize network overhead, we transmit only the latent representation of generated videos, decoding them into pixel-space frames on demand within the service. 

The reward service is architected as a high-throughput, asynchronous, overlapping pipeline comprising two logically separated stages: Decoding and Reward Calculation. The Decoding stage reconstructs video frames from incoming latents, while the Reward Calculation stage concurrently executes multiple reward models in parallel, each computing distinct metrics on the decoded frames. These stages run as independent processes, enabling heterogeneous software environments (e.g., varying PyTorch/CUDA versions or Python dependencies) to coexist without interference. All processes—including one decoder and multiple reward evaluators—run in parallel to maximize throughput. Such a design also ensures high scalability and flexibility. Inter-stage data transfer leverages CUDA Inter-Process Communication (IPC) for zero-copy sharing of decoded frames between workers, eliminating costly host-device memory copies. When resources are disaggregated, NCCL with high bandwidth is employed for efficient inter-stage communication.

\begin{figure}
    \centering
    \includegraphics[width=0.5\linewidth]{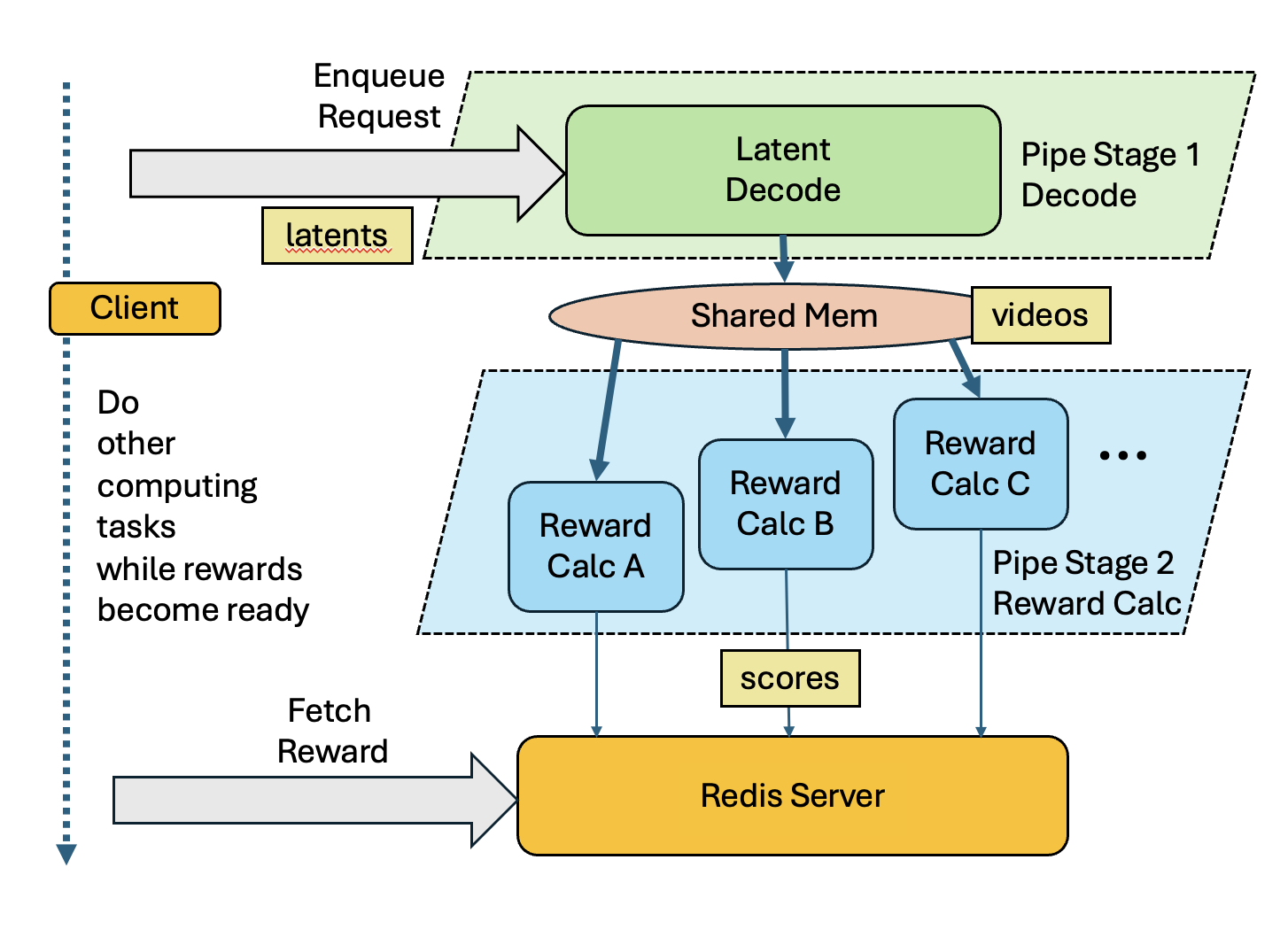}
    \caption{An illustration of the reward server infrastructure.}
    \label{fig:placeholder}
\end{figure}

The pipeline processes multiple videos concurrently in a producer-consumer fashion, overlapping decoding and inference to fully utilize GPU and CPU resources. Each reward request is assigned a unique UUID upon enqueue and returns immediately, enabling asynchronous execution. Computed rewards are persisted in a Redis-backed key-value store, indexed by UUID, and retrieved on demand. During the interval between enqueue and result fetch, the calling system may proceed with other tasks to maximize utilization. The system natively supports batched inference: multiple videos can be submitted in a single request, each evaluated independently across all configured reward models. This architecture enables efficient resource utilization, seamless scalability across multiple GPUs and nodes, and robust fault isolation—critical for large-scale video generation evaluation at scale.

\subsection{Image post-training}

\begin{figure}[t]

	\centering
	\begin{minipage}{.18\linewidth}
		\centering
			\includegraphics[width=\linewidth]{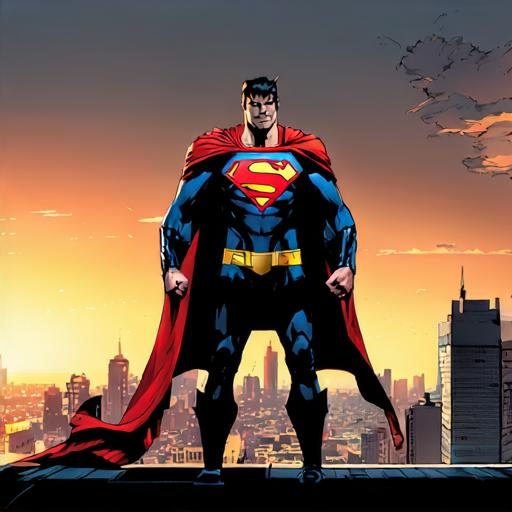}\\
	\end{minipage}
	\begin{minipage}{.18\linewidth}
	\centering
	\includegraphics[width=\linewidth]{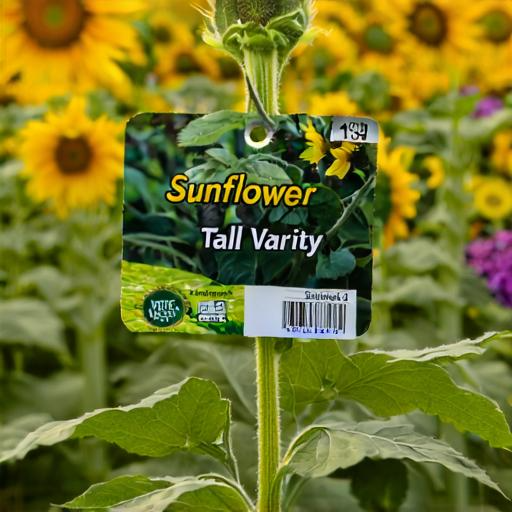}\\
\end{minipage}
	\begin{minipage}{.18\linewidth}
	\centering
	\includegraphics[width=\linewidth]{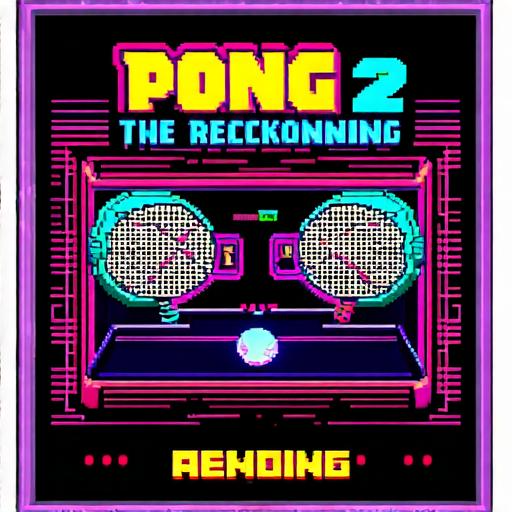}\\
\end{minipage}
	\begin{minipage}{.18\linewidth}
	\centering
	\includegraphics[width=\linewidth]{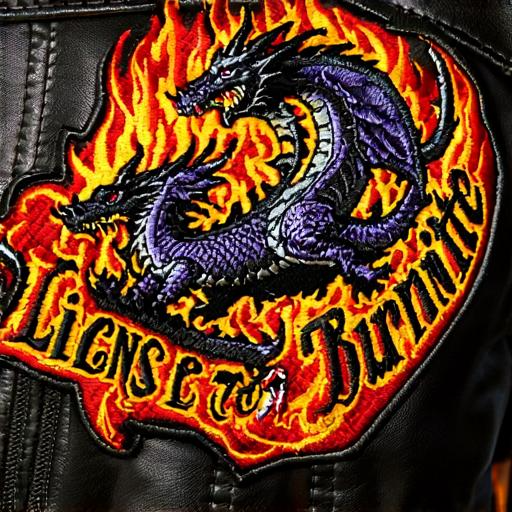}\\
\end{minipage}
	\begin{minipage}{.18\linewidth}
	\centering
	\includegraphics[width=\linewidth]{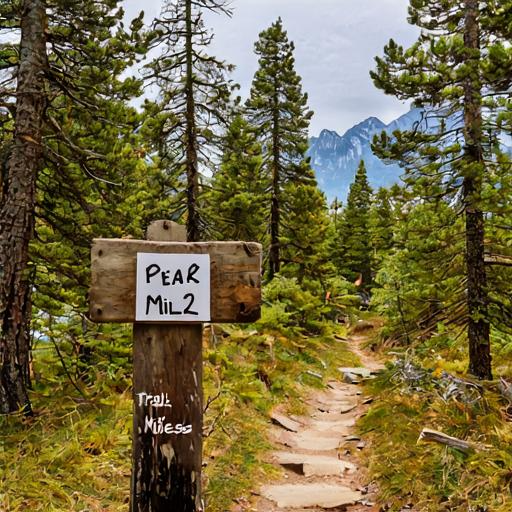}\\
\end{minipage}
   \vspace{-.05in}
	\caption{\label{fig:ref-images} Examples of reference images used for diffusion loss regularization in the absence of real data.}
	\vspace{-.15in}
\end{figure}

\paragraph{Model, reward and data.} We apply \alg to text-to-image (T2I) tasks based on $512\times512$ resolution SD3.5-Medium~\citep{esser2024scaling}, with settings aligned with FlowGRPO~\citep{liu2025flow}. We focus on improving the OCR reward for visual text rendering, which is the most vulnerable to reward hacking among FlowGRPO's settings. We adopt the same text accuracy calculation strategy as FlowGRPO based on edit distance. We use synthetic data sampled from the base model with 40 ODE steps and its default CFG scale of 4.5 for diffusion loss regularization. The synthetic data are visualized in Figure~\ref{fig:ref-images}. Despite they are not accurate in text rendering, they provide abundant information of background details to regularize the RL process.

\paragraph{Training and evaluation configurations.} Our setup largely follows FlowGRPO, adopting the same number of groups per epoch (48), group size (24), LoRA configuration ($\alpha=64, r=32$), and its training and evaluation prompts. We decrease the learning rate from $3e-4$ to $1e-4$ as we find it more stable for DDRL. The relative weight of the diffusion loss is set to 0.001. For synthetic image data, forward noising and diffusion loss computation are performed exactly on the corresponding sampling timesteps. We employ a 10-step SDE sampler for FlowGRPO training and a 40-step ODE sampler for evaluation. The DanceGRPO baseline is implemented by modifying FlowGRPO to remove the reverse KL regularization and use the same initial noise for data collection of the same prompt. The models are trained from around 300 iterations, and we choose DDRL and baseline checkpoints with similar final OCR accuracy for evaluation.
\section{Results Breakdown}

We present a breakdown of \cref{tab:main_table} in \cref{tab:main_results}. This table is also a comprehensive result of \cref{fig:hack_radar}. Notice that for each algorithm, the table contain results of four post-trained models (two model size and two rewards). We mix T2V and I2V data during training, and the resulting checkpoint is used both for T2V and I2V evaluation.

\begin{table*}[t]
\centering
\caption{Breakdown of the reward scores. We compare the base model against DanceGRPO, FlowGRPO, and DDRL across Text-to-Video (T2V) and Image-to-Video (I2V) settings.}
\label{tab:main_results}
\vspace{-10pt}
\resizebox{\textwidth}{!}{%
\begin{tabular}{@{}lllcccccccccccc@{}}
\toprule
& & & \multicolumn{5}{c}{VideoAlign (3 scores)} & \multicolumn{7}{c}{VBench (5 scores)} \\
\cmidrule(lr){4-8} \cmidrule(lr){9-15}
\textbf{Model} & \textbf{Setting} & \textbf{Method} & \textbf{text} & \textbf{motion} & \textbf{visual} & & & \textbf{subject} & \textbf{background} & \textbf{motion} & \textbf{aesthetic} & \textbf{imaging} & & \\
& & & \textbf{alignment} & \textbf{quality} & \textbf{quality} & \textbf{Mean} & \textbf{STD} & \textbf{consistency} & \textbf{consistency} & \textbf{smoothness} & \textbf{quality} & \textbf{quality} & \textbf{Mean} & \textbf{STD} \\
\midrule
\multirow{8}{*}{Cosmos2.5-2B} & \multirow{4}{*}{T2V} 
  & w/o RL & 1.69 & -0.45 & -0.01 & 0.408 & 0.520 & 0.94 & 0.95 & 0.99 & 0.54 & 0.73 & 0.830 & 0.027 \\
& & DanceGRPO & 1.50 & 0.16 & 0.49 & 0.715 & 0.683 & 0.97 & 0.97 & 0.99 & 0.57 & 0.75 & 0.849 & 0.022 \\
& & FlowGRPO & 1.67 & -0.43 & -0.02 & 0.408 & 0.556 & 0.95 & 0.95 & 0.99 & 0.54 & 0.72 & 0.830 & 0.028 \\
& & DDRL & 1.76 & -0.19 & 0.24 & 0.604 & 0.594 & 0.96 & 0.96 & 0.99 & 0.56 & 0.73 & 0.842 & 0.024 \\
\cmidrule{2-15}
 & \multirow{4}{*}{I2V} 
  & w/o RL & 1.57 & -0.82 & -0.52 & 0.079 & 0.434 & 0.92 & 0.94 & 0.99 & 0.52 & 0.69 & 4.050 & 0.160 \\
& & DanceGRPO & 1.46 & -0.45 & -0.24 & 0.254 & 0.534 & 0.94 & 0.94 & 0.99 & 0.53 & 0.71 & 0.822 & 0.028 \\
& & FlowGRPO & 1.57 & -0.81 & -0.55 & 0.069 & 0.436 & 0.92 & 0.94 & 0.99 & 0.51 & 0.67 & 0.807 & 0.033 \\
& & DDRL & 1.61 & -0.66 & -0.42 & 0.177 & 0.465 & 0.94 & 0.95 & 0.99 & 0.52 & 0.70 & 0.819 & 0.030 \\
\midrule
\multirow{8}{*}{Cosmos2.5-14B} & \multirow{4}{*}{T2V} 
  & w/o RL & 1.58 & -0.46 & -0.04 & 0.359 & 0.506 & 0.94 & 0.95 & 0.99 & 0.52 & 0.70 & 0.820 & 0.034 \\
& & DanceGRPO & 1.47 & -0.19 & 0.20 & 0.494 & 0.587 & 0.95 & 0.96 & 0.99 & 0.54 & 0.73 & 0.834 & 0.028 \\
& & FlowGRPO & 1.72 & -0.35 & 0.06 & 0.476 & 0.554 & 0.96 & 0.96 & 0.99 & 0.54 & 0.70 & 0.829 & 0.031 \\
& & DDRL & 1.72 & -0.22 & 0.16 & 0.555 & 0.546 & 0.95 & 0.96 & 0.99 & 0.55 & 0.72 & 0.835 & 0.030 \\
\cmidrule{2-15}
 & \multirow{4}{*}{I2V} 
  & w/o RL & 1.46 & -0.79 & -0.50 & 0.058 & 0.429 & 0.93 & 0.95 & 0.99 & 0.51 & 0.68 & 0.813 & 0.032 \\
& & DanceGRPO & 1.38 & -0.54 & -0.36 & 0.160 & 0.506 & 0.93 & 0.95 & 0.99 & 0.52 & 0.70 & 0.818 & 0.029 \\
& & FlowGRPO & 1.55 & -0.69 & -0.47 & 0.128 & 0.467 & 0.94 & 0.95 & 0.99 & 0.51 & 0.65 & 0.808 & 0.035 \\
& & DDRL & 1.51 & -0.67 & -0.44 & 0.134 & 0.453 & 0.94 & 0.95 & 0.99 & 0.51 & 0.69 & 0.819 & 0.031 \\
\bottomrule
\end{tabular}%
}
\end{table*}

\end{document}